\pdfoutput=1
\documentclass[letterpaper]{article} 
\usepackage{tabularx} 
\usepackage{aaai2026}  
\usepackage{times}  
\usepackage{helvet}  
\usepackage{courier}  
\usepackage[hyphens]{url}  
\usepackage{graphicx} 
\usepackage{natbib}  
\usepackage{caption} 
\usepackage{algorithm}
\usepackage{algorithmic}
\usepackage{graphicx}
\usepackage{subcaption}
\usepackage{float}

\usepackage{newfloat}
\usepackage{listings}
\DeclareCaptionStyle{ruled}{labelfont=normalfont,labelsep=colon,strut=off} 
\floatstyle{ruled}
\newfloat{listing}{tb}{lst}{}
\floatname{listing}{Listing}
\title{Hero-Mamba: Mamba-based Dual Domain Learning for Underwater Image Enhancement}
\author {
    Tejeswar Pokuri \equalcontrib,
    Shivarth Rai \equalcontrib
}
\affiliations {
    Manipal Institute of Technology, Manipal Academy of Higher Education, Manipal - 576104, India \\
    tejeswar.mitmpl2022@learner.manipal.edu, shivarth.mitmpl2022@learner.manipal.edu
}

\begin{document}

\maketitle

\begin{abstract}
Underwater images often suffer from severe degradation, such as color distortion, low contrast, and blurred details, due to light absorption and scattering in water. While learning-based methods like CNNs and Transformers have shown promise, they face critical limitations: CNNs struggle to model the long-range dependencies needed for non-uniform degradation, and Transformers incur quadratic computational complexity, making them inefficient for high-resolution images. To address these challenges, we propose Hero-Mamba, a novel Mamba-based network that achieves efficient dual-domain learning for underwater image enhancement. Our approach uniquely processes information from both the spatial domain (RGB image) and the spectral domain (FFT components) in parallel. This dual-domain input allows the network to decouple degradation factors, separating color/brightness information from texture/noise. The core of our network utilizes Mamba-based SS2D blocks to capture global receptive fields and long-range dependencies with linear complexity, overcoming the limitations of both CNNs and Transformers. Furthermore, we introduce a  ColorFusion block, guided by a background light prior, to restore color information with high fidelity. Extensive experiments on the LSUI and UIEB benchmark datasets demonstrate that Hero-Mamba outperforms state-of-the-art methods. Notably, our model achieves a PSNR of 25.802 and an SSIM of 0.913 on LSUI, validating its superior performance and generalization capabilities.
\end{abstract}


\section{Introduction}
Underwater imaging technology is a critical tool for studying and exploring the Earth's oceans and seas, providing essential data for understanding marine ecosystems, climate change, and human impact \cite{intro1}. It plays a crucial role in a wide range of applications, including autonomous underwater vehicle (AUV) navigation, marine biology, seabed mapping, and underwater structure inspection. The purpose of Underwater Image Enhancement (UIE) is to improve the visual quality of this imagery, a step that is vital for the accuracy of these downstream computer vision tasks and for our scientific understanding of the underwater world.

The quality of underwater images, however, is often severely compromised by the unique optical properties of water \cite{intro2}. The medium itself, along with dissolved impurities and suspended particles,
\begin{figure}[t]
\centering
\includegraphics[width=1.0\columnwidth]{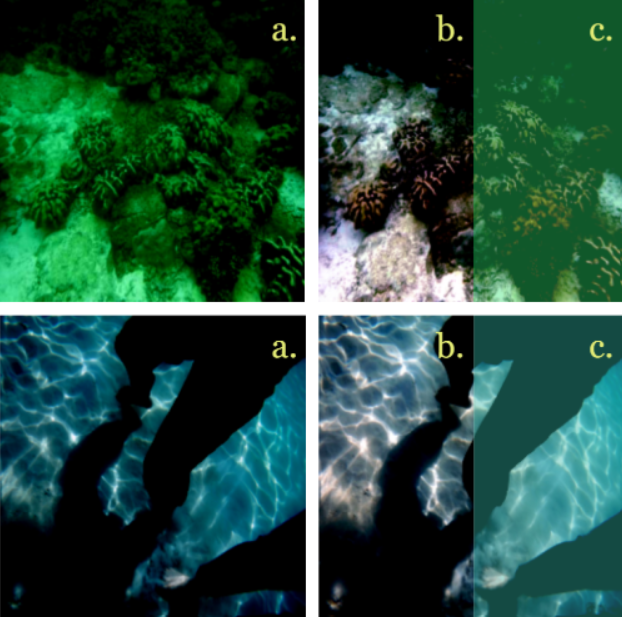} 
\caption{Visualizing the contribution of the background light prior on UIEB samples. (a) Original degraded images. (b) Ground Truth. (c) The background light prior, as estimated by our ColorFusion block, overlaid for visualization. The stark color difference between (b) and (c) highlights the severe color degradation caused by background light.}
\label{fig:fc4} 
\end{figure}
causes strong light absorption and scattering. This degradation manifests as low contrast, blurred details, and significant color distortion. Light absorption is wavelength-dependent, with longer wavelengths (like red) attenuating much faster than shorter wavelengths (like blue and green). This phenomenon is the primary cause of the characteristic blue-green color cast seen in most underwater photographs, which obscures the true colors of the scene \cite{intro3}.

For a mathematical description of this degradation, we can utilize the simplified underwater imaging model provided by \cite{mcglamery} and \cite{jaffe}. This model postulates that a degraded image $I$ acquired by a camera comprises two primary constituents: direct illumination and backscattering.The direct illumination is the true scene radiance $J$ (the clear image) attenuated by the medium transmission rate, $t$. The backscattering is the ambient background light $B$ scattered by particles into the camera's path, veiling the scene. This relationship can be expressed as:
\begin{equation}
I_c(x) = J_c(x)t_c(x) + B_c(1 - t_c(x)),
\label{eq:1}
\end{equation}
here, $I_c(x)$ is the observed degraded image pixel, $J_c(x)$ is the clear image, $t_c(x)$ is the transmission map, and $B_c$ is the background light.

Traditional physical-model based UIE methods \cite{prior1,prior2,prior3,prior4} aim to directly estimate the parameters of the imaging model to reverse the degradation. However, these models require manual tuning of parameters that perform sub optimally in diverse underwater environments leading to poor generalization. With the rise of deep learning, Convolutional Neural Networks (CNNs) became a dominant approach \cite{cnn1,cnn2,cnn3,cnn4}. CNNs can automatically learn complex feature representations from large datasets for end-to-end image restoration. However, their reliance on static, local convolutional kernels limits their receptive field, making it difficult to capture long-range pixel dependencies required to accurately non-uniform degradation across an entire image.

Recently, Transformers and their self-attention mechanism have been extended to vision tasks with great success \cite{Ushape,transformer2,transformer3,transformer4}. Using self-attention mechanisms, Transformers can capture global context and model long-range dependencies effectively. However, their primary drawback is the quadratic computational complexity with respect to the image's spatial size, making them computationally expensive and difficult to apply to high-resolution underwater images.

To address the quadratic complexity issue of Transformers,  state space models \cite{ssm} with selective mechanisms such as Mamba have been developed. Mamba \cite{mamba} provides the ability to capture long-range dependencies and global features with linear complexity. These models have gained widespread attention and application in visual tasks such as semantic segmentation and object detection.

In this study, we wish to address the aforementioned complexities of underwater image enhancement, and make the following contributions:  
\begin{itemize}
    \item We propose Hero-Mamba, a novel U-shaped network that introduces a Mamba-based dual-domain learning paradigm. By processing spatial (RGB) and spectral (FFT) features in parallel, our model efficiently captures global long-range dependencies with linear complexity while simultaneously decoupling degradation factors for more effective enhancement.
    \item Leveraging a physical imaging model-based ColorFusion block, we incorporate background light prior in our network for accurate color restoration in underwater scenes.
    \item Extensive quantitative and qualitative experiments demonstrate that Hero-Mamba surpasses state-of-the-art methods on the public UIEB and LSUI benchmark datasets, showing improvements in both structural similarity and perceptual quality.
\end{itemize}

\begin{figure*}[!ht] 
    \centering
    \includegraphics[width=\textwidth]{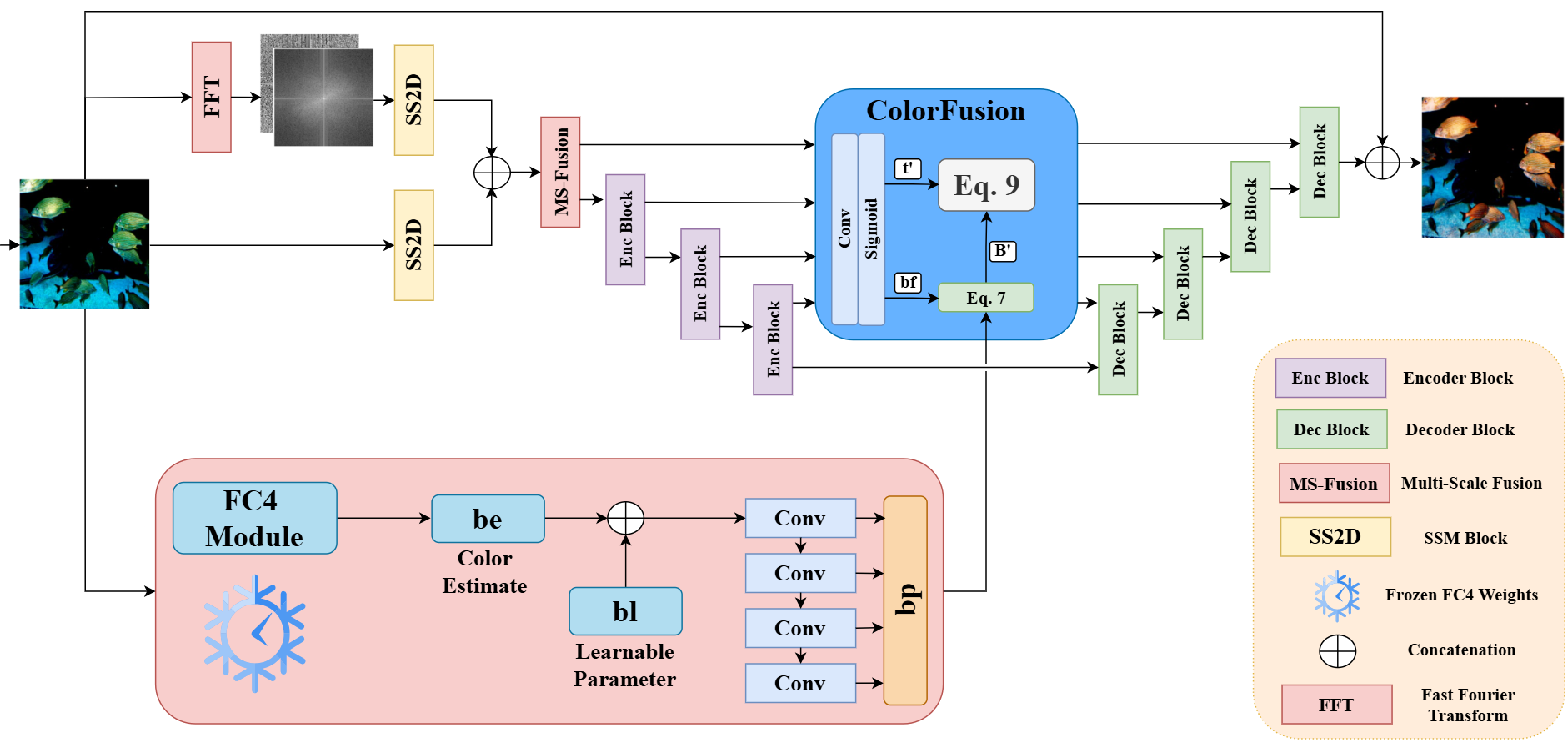}
    \caption{Architectural design of Hero-Mamba, utilizing spatial and spectral domains for accurate feature reconstruction, and ColorFusion block for enhanced color restoration. Using SS2D layers allows for long-range feature learning efficiently.}
    \label{fig:main}
\end{figure*}

\section{Related Work}
\textbf{Traditional methods.} \cite{prior1} presents a hybrid method for contrast enhancement using gamma correction and probability distribution of luminance pixels, to effectively enhance dimmed images by balancing high visual quality with low computational cost. \cite{prior2} introduces the underwater dark channel prior (UDCP), an adaptation of the dark channel prior for estimation of transmission in underwater scenes. \cite{prior4} account for the wavelet-dependent attenuation of light in underwater environments, proposing a physics-based model for image restoration utilizing haze-lines prior. 

\textbf{CNN-based methods.} \cite{cnn1} introduces a novel semi-supervised learning framework. Combining a reliable pseudo-labeling mechanism with contrastive learning, the framework effectively exploits unlabeled data to improve model performance on underwater images. \cite{cnn3} introduces Ucolor, a network combining multi-color space embedding with physics-inspired guidance. The network processes and adaptively fuses features from RGB, Lab and HSV inputs to capture characteristics from different color spaces. Using physical imaging model, the network computes a medium transmission map, guiding it to focus on regions with higher degradation. \cite{cnn4} resolves UIE into distribution estimation and consensus process. Combining a conditional variational autoencoder (CVAE) with adaptive instance normalization (AdaIN), the network learns to model a distribution of possible enhanced outputs for a single input. Following this, a consensus process is utilized to predict a deterministic result from the distribution. 

\textbf{Transformer-based methods.} \cite{Ushape} introduces the U-Shape Transformer, integrated with a channel-wise multi-scale feature fusion transformer (CMSFFT) module and a spatial-wise global feature modeling transformer (SGFMT). These modules, designed specifically for UIE, target inconsistent attenuation across channels and spatial areas. \cite{transformer2} presents Spectroformer, a novel UIE transformer, integrating spatial and frequency domains using its Multi-Domain Query Cascaded Attention (MQCA) mechanism, which uses frequency-domain queries with spatial keys/values. A Spatio-Spectro Fusion Attention Block enhances feature propagation in skip connections by fusing both domains. Finally, a Hybrid Fourier-Spatial Upsampling Block combines upsampling techniques for superior feature resolution enhancement.\cite{transformer4} proposes a U-Net based network employing Swin Transformer blocks. This work introduces Reinforced Swin-Convs Transformer Block (RSCTB), which incorporates convolutions within the Swin Transformer's attention mechanism to capture local attention alongside the Transformer's global dependency modeling.

\textbf{SSM-based methods.} \cite{SSuie} propose SS-UIE, a Mamba-based network achieving spatial-spectral dual-domain adaptive learning with linear complexity, addressing limitations of prior CNN and Transformer methods. It introduces the Spatial-Spectral (SS) block, which combines a Mamba-inspired Multi-scale Cycle Selective Scan (MCSS) for global spatial modeling and an FFT-based Spectral-Wise Self-Attention (SWSA) for spectral modeling in parallel, allowing the network to model degradations of different spatial regions and spectral bands. \cite{mamba-uie} introduces Mamba-UIE, a physical model constraint-based UIE framework. The network utilizes Mamba-In-Convolution Blocks (MIC) to capture long-range dependencies in the spatial and channel dimensions, while the CNN backbone extracts local features. \cite{water-mamba} introduce WaterMamba, which incorporates the proposed Spatial-Channel Omnidirectional Selective Scan (SCOSS) blocks. SCOSS blocks model pixel-channel dependencies by processing spatial and channel information in four directions.

\section{Methodology}
\subsection{Overall Architecture}
The proposed Hero-Mamba network (Fig.\ref{fig:main}) primarily consists of encoder, decoder and ColorFusion blocks, following a  U-shaped structure with residual connections for multi-scale feature fusion. Given a degraded underwater image $I \in {R}^{3 \times H \times W}$, $I$ is first passed through a two-dimensional selective-scan module (SS2D) to obtain encoded spatial features. Since underwater image degradation causes global losses, utilizing SS2D modules allows us to capture long-range spatial information efficiently. In parallel, we apply the Fast Fourier Transform (FFT) to obtain amplitude component and phase component of $I$, and concatenate them together to obtain $I_S \in {R}^{2 \times H \times W}$, a spectral representation of the input image. As discussed in \cite{frequency}, the amplitude and phase components obtained by applying FFT can allow the network to enhance overall brightness and color accuracy by processing the amplitude while separately enhancing details and reducing noise by processing the phase. $I_S$ is passed through a SS2D layer, the output of which is concatenated to the output of the parallel SS2D with $I$ as input, to get the initial feature representation, $I_F \in {R}^{5 \times H \times W}$. This operation is represented as follows:
\begin{equation}
I_F = SS2D(I) \oplus SS2D(I_S),
\label{eq:2}
\end{equation}
where $\oplus$ represents the concatenation operation. 

\begin{figure}
    \centering
    \includegraphics[width=1.0\linewidth]{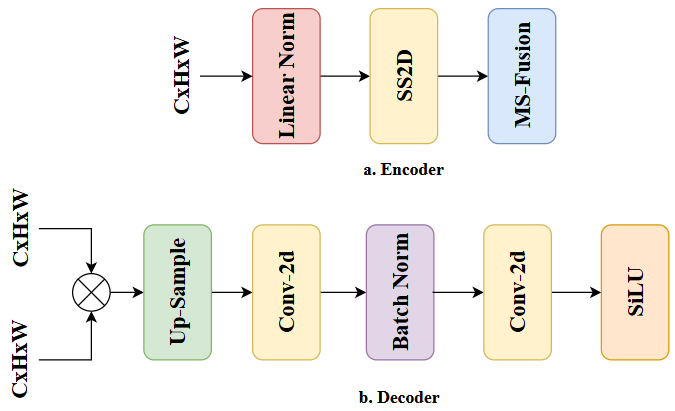}
    \caption{Overview of the Encoder Block}
    \label{fig:encoder}
\end{figure}

Following this, $I_F$ passes through a MS-Fusion block and three encoder blocks, each down samples features by a factor of 2, to dimensions of $f_1 \in {R}^{32 \times\frac{H}{2}\times\frac{W}{2}}$, $f_2 \in {R}^{64 \times\frac{H}{4}\times\frac{W}{4}}$, $f_3 \in {R}^{128 \times\frac{H}{8}\times\frac{W}{8}}$ and $f_4 \in {R}^{256 \times\frac{H}{16}\times\frac{W}{16}}$ respectively. The encoder block, as shown in Fig. \ref{fig:encoder}, a SS2D layer followed by MS-Fusion block, is designed to capture global dependencies in linear complexity and extract multi-scale features, providing a rich feature representation of the input.

Simultaneously, each of the multi-scale features, $f_i \space \forall \space i \in \{1, 2, 3, 4\}$, are fed through a ColorFusion block. In the architecture, we use ColorFusion blocks to preserve the color accuracy of the generated output. They are discussed in the following subsection. Let the output of each ColorFusion block be represented by $c_i \space \forall \space i \in \{1, 2, 3, 4\}$, then $c_i$ can be represented as:
\begin{equation}
c_i = ColorFusion(f_i)
\label{eq:3}
\end{equation}

Then, feature $f_4$ passes through the decoder network, consisting of four decoder blocks. As in Fig. \ref{fig:encoder}, each decoder consists of an up sampling layer, followed by Conv layer, BatchNorm layer, Conv layer and SiLU activation function. Each decoder receives decoded features and ColorFusion features, $c_i$ of corresponding size concatenated together. Let the output of each decoder block be represented by $d_i \space \forall \space i \in \{1, 2, 3, 4\}$, then, in general, $d_i$ can be represented as:
\begin{equation}
d_i = SiLU(Conv(BN(Conv(Up(d_{i-1} \oplus c_i)))))
\label{eq:4}
\end{equation}
where $d_{i-1}$ is the decoded feature from preceding decoder block and $\oplus$ represents the concatenation operation. 


\subsection{ColorFusion Block}
Motivated by the success of Joint Prior Module in \cite{jpm}, we adapt a simplified version in the form of ColorFusion block in our network. As discussed earlier, severe color degradation is prevalent in underwater imagery due to selective absorption of longer wavelength signals cause images to have a predominant blue/green hue. In the underwater imaging model described in Eq. \ref{eq:1}, background light heavily affects color and visibility (see Fig.\ref{fig:fc4}). Thus, using it as "prior knowledge" provides a stable, theory-based lighting model that helps guide the restoration process. 

The blue/green hue for each degraded image can be estimated as background light prior using the theory of color constancy \cite{theory, theory2}. An initial background light prior, $b_e$ is estimated using a FC4 \cite{fc4} model pre-trained on color-constant data \cite{fc4Data}. Since the pre-training data is captured on land, estimated $b_e$ is an approximate, and to further tune the prior for underwater environments, a learnable parameter, $b_l$, is computed along with $b_e$. The resultant estimated background light prior, $b_p$, then, can be expressed as:
\begin{equation}
b_p = Conv(b_e + b_l),
\label{eq:5}
\end{equation}
where $+$ is the expansion operation. Together with $b_p$, a background light feature is extracted directly from the input image features. This feature, $b_f$, is represented as:
\begin{equation}
b_f = Sigmoid(Conv(f_i)), 
\label{eq:6}
\end{equation}
where $f_i$ is the feature extracted in the encoder at corresponding scale. Then, $b_p$ and $b_f$ are dynamically mixed in order to get an accurate background light prior estimate for the input image. This is represented by:
\begin{equation}
B' = \omega b_f + (1-\omega)b_p,
\label{eq:7}
\end{equation}
where $\omega$ is a dynamic weight coefficient ranging from 0 to 1.
Now, using the underwater image model in Eq.\ref{eq:1}, we can approximate \cite{app} the undegraded image $J$ as:
\begin{equation}
J = It + B(1-t) 
\label{eq:8}
\end{equation}
In terms of feature representations in the network, Eq.\ref{eq:8} can be written as:
\begin{equation}
c_i = f_it' + B'(1-t'),
\label{eq:9}
\end{equation}
where $c_i$ is the output of the ColorFusion block, $f_i$ is the encoded input feature, $B'$ is the background light prior, and $t'$ is the feature transmission map, estimated as follows:
\begin{equation}
t' = Sigmoid(Conv(f_i)
\label{eq:10}
\end{equation}

\begin{figure}[!ht]
    \centering
    \includegraphics[width=1\linewidth]{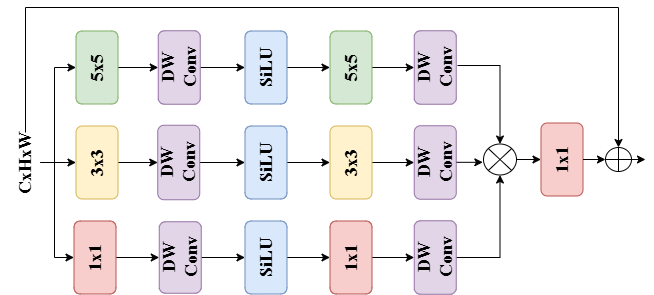}
    \caption{Overview of the MS-fusion block. Parallel branches of $1\times1$, $3\times3$ and $5\times5$ kernel sizes facilitate comprehensive feature extraction.}
    \label{fig:ms}
\end{figure}

\subsection{MS-Fusion Block}
As shown in Fig. \ref{fig:ms}, MS-Fusion block simultaneously processes features at three spatial scales, providing a comprehensive feature representation required for restoring local details and sharp edges. The MS-Fusion block has a multi-branch structure, where the first branch can be defined as:
\begin{equation}
B_1 = D_W^{1\times1}C^{1\times1}(SiLU(D_W^{1\times1}C^{1\times1}(F))),
\label{eq:11}
\end{equation}
where $D_W^{1\times1}$ denotes depthwise convolution of size $1\times1$, $C^{1\times1}$ represents $1\times1$ convolution, and $F$ is the input feature. Similarly, the second and third branches perform the same set of operations with $3\times3$ and $5\times5$ kernel sizes, respectively. Following this, the features from all three branches, $B_1$, $B_2$ and $B_3$, are concatenated and passed through a $1\times1$ convolution layer to reduce feature channels. This feature aggregate is then added back to the input feature to get the MS-Fusion block output. This is represented as:
\begin{equation}
F' = C^{1\times1}(B_1 \oplus B_2 \oplus B_3) + F,
\label{eq:12}
\end{equation}
here $\oplus$ denotes concatenation and, $+$ denotes element-wise addition.

\subsection{Loss Function}
In order to holistically address the multi-faceted nature of degradation in underwater images, such as low contrast, blurred details and color distortion, we use a composite loss function with L1 loss, SSIM loss and contrastive loss components. 

\textbf{L1 Loss:} calculates the average of absolute difference between every single pixel in the model's output image and the corresponding pixel in the ground truth. This component promotes pixel-level accuracy, making sure the colors and brightness values of the enhanced image are as close as possible to the ground truth. L1 loss is defined as follows:
\begin{equation}
\mathcal{L}_1 = \frac{\sum_{i=1}^{n} \left| f(X_i) - Y_i \right|}{n},
\label{eq:13}
\end{equation}
where $f(X_i)$ represents generated output and $Y_i$ is ground truth for the $i^{th}$ sample, and, $n$ is the number of samples.

\begin{table*}[htbp]
\centering
\caption{Quantitative comparison on LSUI Dataset.The best result is
highlighted in red and the second best result is in blue.}
\begin{tabular*}{\textwidth}{@{\extracolsep{\fill}}l l c c c c}
\hline
\textbf{Paper Name} & \textbf{Venue} & \textbf{SSIM} $\uparrow$ & \textbf{PSNR} $\uparrow$ & \textbf{LPIPS} $\downarrow$ & \textbf{FSIM} $\uparrow$ \\
\hline
U-Gan & ICRA 2018 & 0.772 & 19.423 & 0.374 & 0.763 \\
FUnIE-Gan & RAL 2020 & 0.798 & 20.783 & 0.234 & 0.833 \\
U shape Transformer & TIP 2023 & 0.821 & 21.623 & 0.298 & 0.847 \\
SS-UIE & AAAI 2025 & 0.816 & 19.093 & 0.270 & 0.892 \\
CE-VAE & WACV 2024 & 0.832 & 22.638 & {\color{blue}0.127} & 0.932 \\
Water Mamba & Arxiv 2024 & {\color{blue}0.877} & {\color{blue}23.463} & {0.134} & {\color{blue}0.937} \\
\hline
\textbf{Hero-Mamba} (Ours) &  & {\color{red}0.913} & {\color{red}25.802} & {\color{red}0.117} & {\color{red}0.958} \\
\hline
\end{tabular*}
\end{table*}

\begin{table*}[htbp]
\centering
\caption{Quantitative comparison on UIEB Dataset. The best result is
highlighted in red and the second best result is in blue.}
\begin{tabular*}{\textwidth}{@{\extracolsep{\fill}}l l c c c c}
\hline
\textbf{Paper Name} & \textbf{Venue} & \textbf{SSIM} $\uparrow$ & \textbf{PSNR} $\uparrow$ & \textbf{LPIPS} $\downarrow$ & \textbf{FSIM} $\uparrow$ \\
\hline
U-Gan & ICRA 2018 & 0.805 & 19.676 & 0.197 & 0.912 \\
FUnIE-Gan & RAL 2020 & 0.814 & 18.781 & 0.163 & 0.919 \\
Semi-UIR & CVPR 2023 & 0.821 & 23.400 & 0.157 & 0.932 \\
PUIE-Net & ECCV 2022 & 0.854 & 21.501 & 0.132 & 0.863 \\
WF-Diff & CVPR 2024 & 0.873 & {\color{red}27.260} & 0.139 & 0.897 \\
NU2Net & AAAI 2023 & 0.907 & 22.633 & {\color{red}0.100} & 0.949 \\
Water Mamba & Arxiv 2024 & {\color{blue}0.931} & {\color{blue}24.751} & 0.143 & {\color{red}0.973} \\
\hline
\textbf{Hero-Mamba} (Ours) &  & {\color{red}0.942} & 24.526 & {\color{blue}0.125} & {\color{blue}0.945} \\
\hline
\end{tabular*}
\end{table*}

\begin{figure*}[!ht]
    \centering
    \begin{subfigure}{0.12\linewidth}
        \centering
        \includegraphics[width=\linewidth]{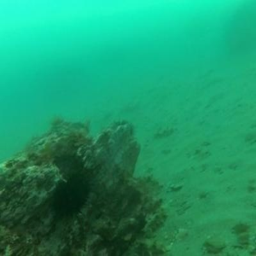}
    \end{subfigure}
    \begin{subfigure}{0.12\linewidth}
        \centering
        \includegraphics[width=\linewidth]{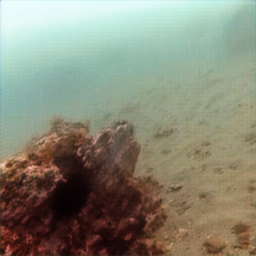}

    \end{subfigure}
    \begin{subfigure}{0.12\linewidth}
        \centering
        \includegraphics[width=\linewidth]{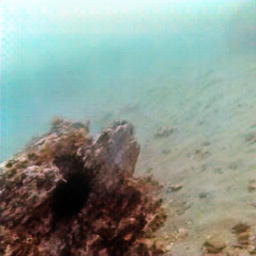}
        
    \end{subfigure}
    \begin{subfigure}{0.12\linewidth}
        \centering
        \includegraphics[width=\linewidth]{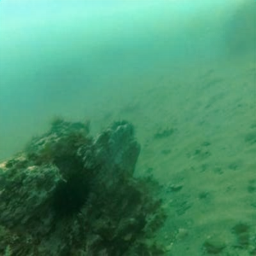}
        
    \end{subfigure}
    \begin{subfigure}{0.12\linewidth}
        \centering
        \includegraphics[width=\linewidth]{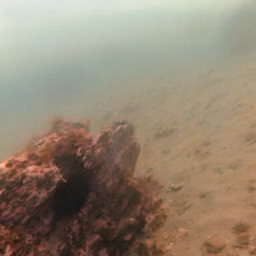}
        
    \end{subfigure}
    \begin{subfigure}{0.12\linewidth}
        \centering
        \includegraphics[width=\linewidth]{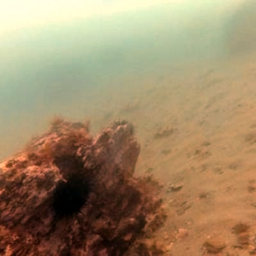}
    \end{subfigure}
    \begin{subfigure}{0.12\linewidth}
        \centering
        \includegraphics[width=\linewidth]{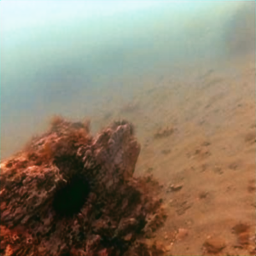}
    \end{subfigure}
    \begin{subfigure}{0.12\linewidth}
        \centering
        \includegraphics[width=\linewidth]{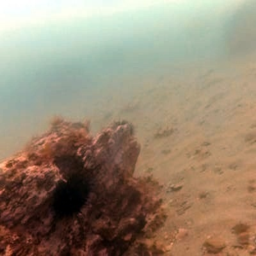}
    \end{subfigure}

    \vspace{0.1cm}

    \begin{subfigure}{0.12\linewidth}
        \centering
        \includegraphics[width=\linewidth]{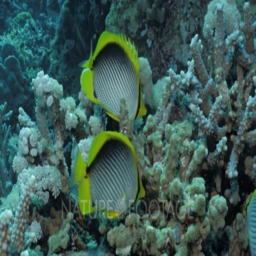}
    \end{subfigure}
    \begin{subfigure}{0.12\linewidth}
        \centering
        \includegraphics[width=\linewidth]{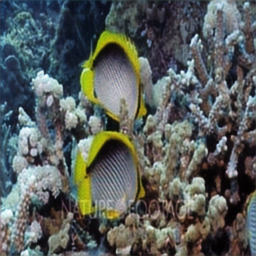}
    \end{subfigure}
    \begin{subfigure}{0.12\linewidth}
        \centering
        \includegraphics[width=\linewidth]{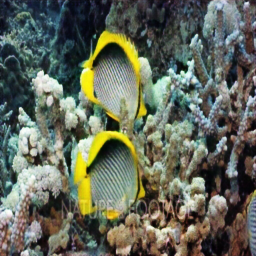}
    \end{subfigure}
    \begin{subfigure}{0.12\linewidth}
        \centering
        \includegraphics[width=\linewidth]{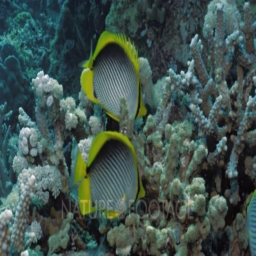}
    \end{subfigure}
    \begin{subfigure}{0.12\linewidth}
        \centering
        \includegraphics[width=\linewidth]{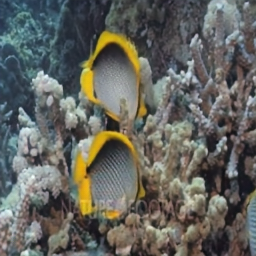}
    \end{subfigure}
    \begin{subfigure}{0.12\linewidth}
        \centering
        \includegraphics[width=\linewidth]{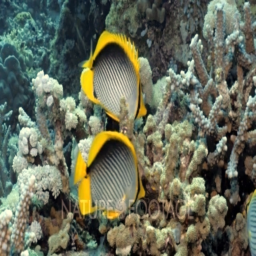}
    \end{subfigure}
    \begin{subfigure}{0.12\linewidth}
        \centering
        \includegraphics[width=\linewidth]{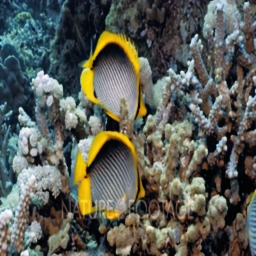}
    \end{subfigure}
    \begin{subfigure}{0.12\linewidth}
        \centering
        \includegraphics[width=\linewidth]{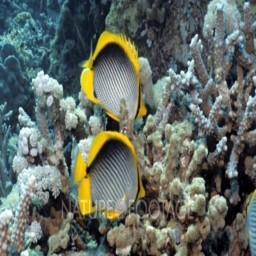}
    \end{subfigure}

    \vspace{0.1cm}

    \begin{subfigure}{0.12\linewidth}
        \centering
        \includegraphics[width=\linewidth]{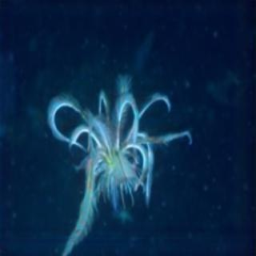}
    \end{subfigure}
    \begin{subfigure}{0.12\linewidth}
        \centering
        \includegraphics[width=\linewidth]{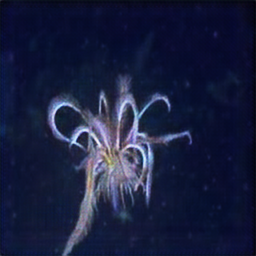}
    \end{subfigure}
    \begin{subfigure}{0.12\linewidth}
        \centering
        \includegraphics[width=\linewidth]{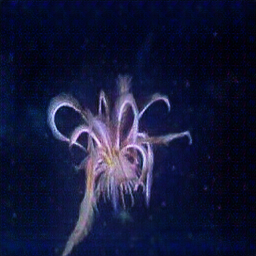}
    \end{subfigure}
    \begin{subfigure}{0.12\linewidth}
        \centering
        \includegraphics[width=\linewidth]{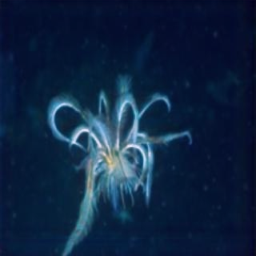}
    \end{subfigure}
    \begin{subfigure}{0.12\linewidth}
        \centering
        \includegraphics[width=\linewidth]{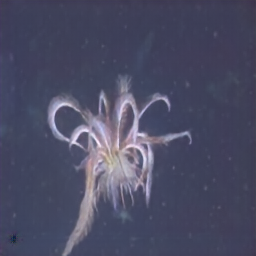}
    \end{subfigure}
    \begin{subfigure}{0.12\linewidth}
        \centering
        \includegraphics[width=\linewidth]{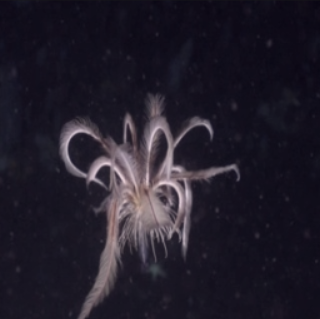}
    \end{subfigure}
    \begin{subfigure}{0.12\linewidth}
        \centering
        \includegraphics[width=\linewidth]{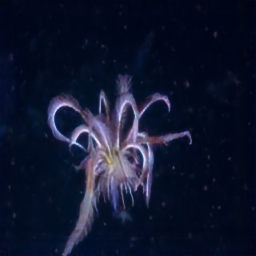}
    \end{subfigure}
    \begin{subfigure}{0.12\linewidth}
        \centering
        \includegraphics[width=\linewidth]{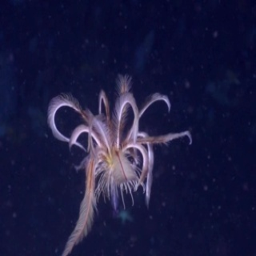}
    \end{subfigure}

    \vspace{0.2cm}
    \begin{minipage}{0.12\linewidth}\centering a) Input \end{minipage}
    \begin{minipage}{0.12\linewidth}\centering b) U-Gan \end{minipage}
    \begin{minipage}{0.12\linewidth}\centering c) FUnIE-Gan \end{minipage}
    \begin{minipage}{0.12\linewidth}\centering d) SS-UIE \end{minipage}
    \begin{minipage}{0.12\linewidth}\centering e) CE-VAE \end{minipage}
    \begin{minipage}{0.12\linewidth}\centering f) Water Mamba \end{minipage}
    \begin{minipage}{0.12\linewidth}\centering g) Hero-Mamba \end{minipage}
    \begin{minipage}{0.12\linewidth}\centering h) Ground Truth \end{minipage}

    \caption{Visual comparison of enhancement results by various models on LSUI dataset.}
    \label{fig:lsui}
\end{figure*}

\textbf{SSIM Loss:} based on the Structural Similarity Index (SSIM), a metric designed to measure the perceptual quality of an image, which aligns better with human judgment. SSIM Loss is defined as follows:
\begin{equation}
\mathcal{L}_{ssim} = 1 - \frac{(2\mu_X \mu_Y + \varepsilon_1) (2\sigma_{XY} + \varepsilon_2)}{(\mu_X^2 + \mu_Y^2 + \varepsilon_1) (\sigma_X^2 + \sigma_Y^2 + \varepsilon_2)},
\label{eg:14}
\end{equation}
where $\mu_X$ and $\mu_Y$ represent the respective means of $X$ and $Y$, while $\sigma_X^2$ and $\sigma_Y^2$ are their respective variances. $\sigma_{XY}$ denotes the covariance between $X$ and $Y$. $\varepsilon_1$ and $\varepsilon_2$ are small constants included for numerical stability.

\begin{table*}[htbp]
\centering
\caption{Cross-dataset study results for Hero-Mamba.}
\begin{tabular*}{\textwidth}{@{\extracolsep{\fill}}l l c c c c} 
\hline
\textbf{Train Dataset} & \textbf{Test Dataset} & \textbf{SSIM} $\uparrow$ & \textbf{PSNR} $\uparrow$ & \textbf{FSIM} $\uparrow$ & \textbf{LPIPS} $\downarrow$ \\
\hline
LSUI & UIEB & 0.868 & 19.655 & 0.934 & 0.129 \\
UIEB & LSUI & 0.846 & 20.380 & 0.919 & 0.206 \\
\hline
\end{tabular*}
\end{table*}

\begin{figure*}[htbp]
    \centering
    \begin{subfigure}{0.12\linewidth}\centering
        \includegraphics[width=\linewidth]{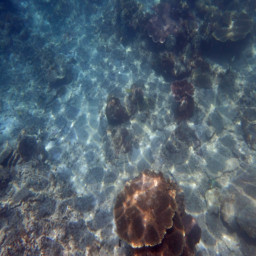}
    \end{subfigure}
    \begin{subfigure}{0.12\linewidth}\centering
        \includegraphics[width=\linewidth]{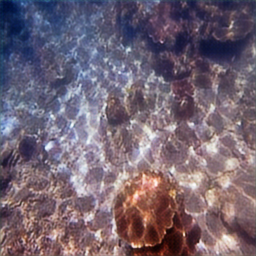}
    \end{subfigure}
    \begin{subfigure}{0.12\linewidth}\centering
        \includegraphics[width=\linewidth]{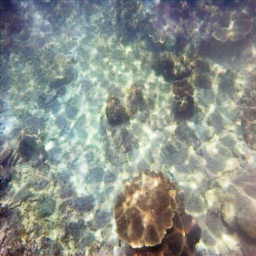}
    \end{subfigure}
    \begin{subfigure}{0.12\linewidth}\centering
        \includegraphics[width=\linewidth]{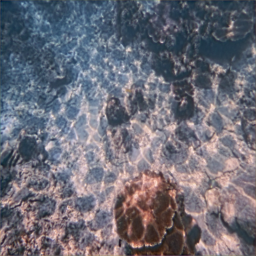}
    \end{subfigure}
    \begin{subfigure}{0.12\linewidth}\centering
        \includegraphics[width=\linewidth]{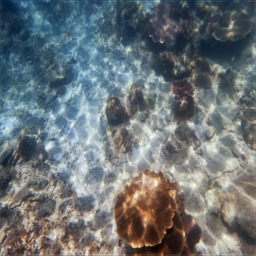}
    \end{subfigure}
    \begin{subfigure}{0.12\linewidth}\centering
        \includegraphics[width=\linewidth]{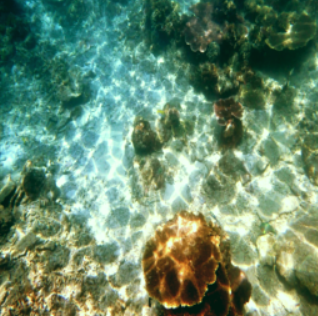}
    \end{subfigure}
    \begin{subfigure}{0.12\linewidth}\centering
        \includegraphics[width=\linewidth]{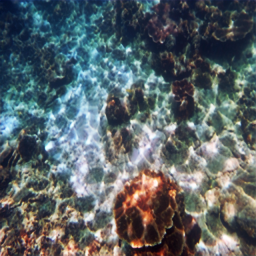}
    \end{subfigure}
    \begin{subfigure}{0.12\linewidth}\centering
        \includegraphics[width=\linewidth]{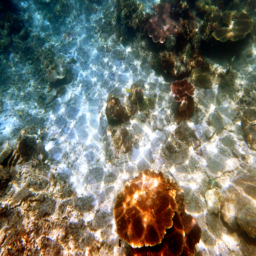}
    \end{subfigure}

    \vspace{0.1cm}
    \begin{subfigure}{0.12\linewidth}\centering
        \includegraphics[width=\linewidth]{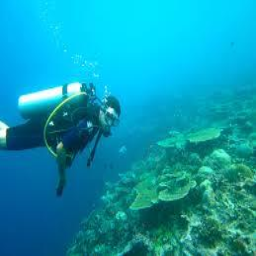}
    \end{subfigure}
    \begin{subfigure}{0.12\linewidth}\centering
        \includegraphics[width=\linewidth]{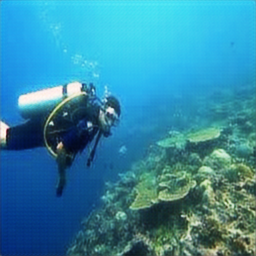}
    \end{subfigure}
    \begin{subfigure}{0.12\linewidth}\centering
        \includegraphics[width=\linewidth]{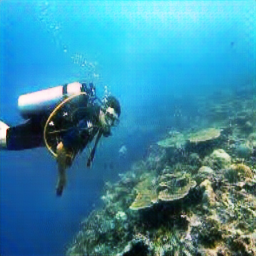}
    \end{subfigure}
    \begin{subfigure}{0.12\linewidth}\centering
        \includegraphics[width=\linewidth]{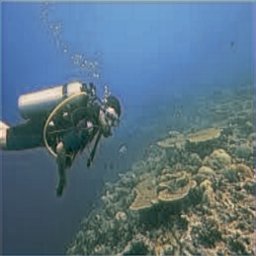}
    \end{subfigure}
    \begin{subfigure}{0.12\linewidth}\centering
        \includegraphics[width=\linewidth]{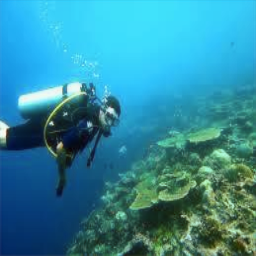}
    \end{subfigure}
    \begin{subfigure}{0.12\linewidth}\centering
        \includegraphics[width=\linewidth]{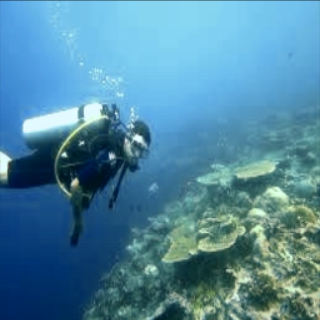}
    \end{subfigure}
    \begin{subfigure}{0.12\linewidth}\centering
        \includegraphics[width=\linewidth]{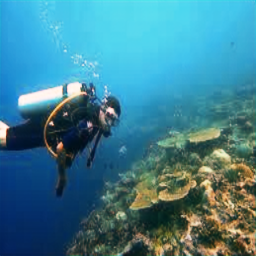}
    \end{subfigure}
    \begin{subfigure}{0.12\linewidth}\centering
        \includegraphics[width=\linewidth]{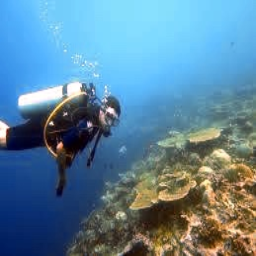}
    \end{subfigure}

    \vspace{0.1cm}
    \begin{subfigure}{0.12\linewidth}\centering
        \includegraphics[width=\linewidth]{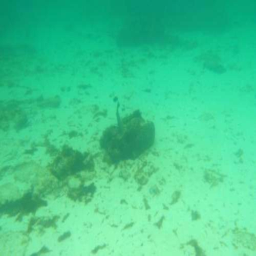}
    \end{subfigure}
    \begin{subfigure}{0.12\linewidth}\centering
        \includegraphics[width=\linewidth]{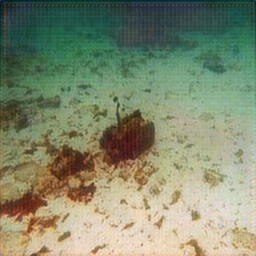}
    \end{subfigure}
    \begin{subfigure}{0.12\linewidth}\centering
        \includegraphics[width=\linewidth]{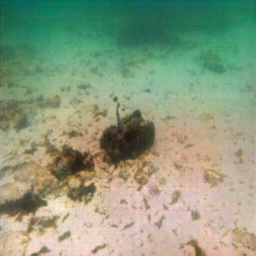}
    \end{subfigure}
    \begin{subfigure}{0.12\linewidth}\centering
        \includegraphics[width=\linewidth]{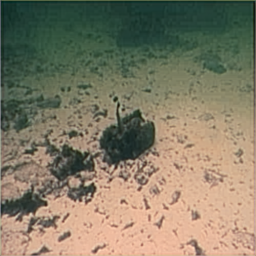}
    \end{subfigure}
    \begin{subfigure}{0.12\linewidth}\centering
        \includegraphics[width=\linewidth]{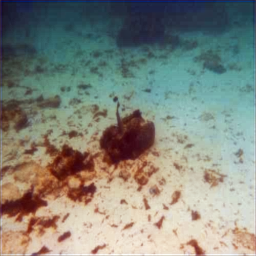}
    \end{subfigure}
    \begin{subfigure}{0.12\linewidth}\centering
        \includegraphics[width=\linewidth]{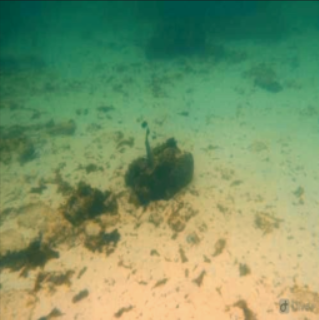}
    \end{subfigure}
    \begin{subfigure}{0.12\linewidth}\centering
        \includegraphics[width=\linewidth]{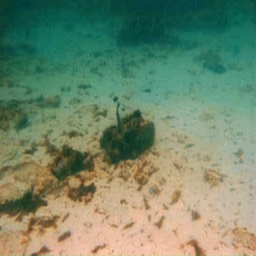}
    \end{subfigure}
    \begin{subfigure}{0.12\linewidth}\centering
        \includegraphics[width=\linewidth]{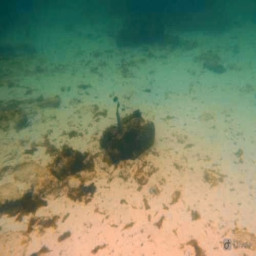}
    \end{subfigure}
    \vspace{0.2cm}
    \begin{minipage}{0.12\linewidth}\centering a) Input \end{minipage}
    \begin{minipage}{0.12\linewidth}\centering b) U-Gan \end{minipage}
    \begin{minipage}{0.12\linewidth}\centering c) FUnIE-Gan \end{minipage}
    \begin{minipage}{0.12\linewidth}\centering d) Semi-UIR \end{minipage}
    \begin{minipage}{0.12\linewidth}\centering f) NU2Net \end{minipage}
    \begin{minipage}{0.12\linewidth}\centering g) Water Mamba \end{minipage}
    \begin{minipage}{0.12\linewidth}\centering h) Hero-Mamba \end{minipage}
    \begin{minipage}{0.12\linewidth}\centering i) Ground Truth \end{minipage}
    \caption{Visual comparison of enhancement results by various methods on UIEB dataset.}
    \label{fig:uieb}
\end{figure*}

\textbf{Contrastive Loss:} helps the model learn a better feature representation, ensuring that the features of the generated image are mathematically closer to the features of the ground truth than to the degraded image. This teaches the model to learn the features of a high-quality underwater image, leading to a more robust and perceptually accurate restoration. This loss is defined as:
\begin{equation}
\min \|J - \phi(I, w)\| + \beta \cdot \rho(\varphi(p), \varphi(n), \varphi(\phi(I, w))),
\label{eq:15}
\end{equation}
where $\min$ represents the minimization objective, $J$ is the reference image, and $\phi(I, w)$ is the predicted undegraded image $J'$, derived through network parameters $w$ from the degraded image $I$, i.e., the anchor sample. $\left\|J - \phi(I, w)\right\|$ measures the difference between $J$ and $J'$, using
the L1 norm. In the second term, $\beta$ is used to adjust the weight of different features of the samples, and $\rho$ measures the similarity of features between samples, also using the $\mathcal{L}_1$ norm. $\phi(p)$, $\phi(n)$, and $\phi(\phi(I, w))$ represent the common intermediate features extracted for all samples through the same pre-trained model, where $p$ is the positive sample, $n$ is the negative sample, and $\phi(I, w)$ is the prediction from the anchor sample.

Our overall composite loss function is expressed as:
\begin{equation}
\mathcal{L} = \alpha \mathcal{L}_{1} + \beta \mathcal{L}_{ssim} + \gamma \mathcal{L}_{contrastive},
\label{eq:16}
\end{equation}
where $\alpha$, $\beta$ and $\gamma$ are the weighting coefficients. After extensive experimentation, they were set to $0.3$, $0.8$ and $0.1$ respectively.


\section{Experiments}
\subsection{Experimental Setup}
\textbf{Implementation Details.} The proposed model was implemented using the PyTorch 2.4.1 framework and was trained and tested on an NVIDIA Tesla T4 GPU. The network was trained end-to-end using the AdamW \cite{adamW} optimizer with a learning rate of $3 \times 10^{-4}$ and $\beta_1 = 0.9$, $\beta_2 = 0.999$. All images were resized to a fixed size of $256 \times 256$ pixels. A training batch size of 4 was used, and the network was trained for 250 epochs. The learning rate was dynamically adjusted using a cosine annealing strategy \cite{cosine}.


\textbf{Datasets.} We utilize two publicly available and widely bench marked underwater image datasets, UIEB \cite{UIEB} and LSUI \cite{Ushape}, for our study. UIEB (Underwater Image Enhancement Benchmark) dataset contains 890 real underwater images with reference. We divide the 890 paired images into a training set of 800 image pairs and validation set of 90 image pairs. LSUI (Large-Scale Underwater Image) dataset contains 4279 real underwater
image pairs. We partition 3851 pairs for the training set and 428 pairs for the test set. These datasets cover various underwater scenes and imaging conditions, such as coral reefs, underwater organisms, and underwater terrains.

\textbf{Evaluation Metrics.} For a comprehensive assessment of the performance of our proposed method, we utilize SSIM \cite{ssim},PSNR \cite{PSNR}, FSIM \cite{fsim} and LPIPS \cite{lpips} metrics. Peak Signal to Noise Ratio (PSNR) computes the ratio of image signal to noise and reflects the overall quality of the generated image. Structural Similarity Index Measure (SSIM) assesses the perceived similarity between two images based on structural information. Feature Similarity Index Measure (FSIM) assesses image quality by comparing low-level features between two images. Learned Perceptual Image Patch Similarity (LPIPS) evaluates the perceptual similarity between two image patches using features learned by deep neural networks (eg. VGG19).

\begin{table}[htbp]
\centering
\caption{Break-down ablation study for Hero-Mamba.}
\begin{tabular}{lcc}
\hline
\textbf{Model} & \textbf{SSIM}\\
\hline
Base & 0.847\\
Base + MS-Fusion & 0.872\\
Base + MS-Fusion + SS2D & 0.890\\
Base + MS-Fusion + SS2D + FFT & 0.914\\
Base + MS-Fusion + SS2D + FFT + ColorFusion & 0.942\\
\hline
\end{tabular}
\label{tab:abalation}
\end{table}
\textbf{Comparison Methods.} To validate the performance of our proposed method, we perform a comparative analysis between our Hero-Mamba and 10 SOTA methods for UIE. 
These methods include U-GAN \cite{UGan}, FUnIE-GAN \cite{FUnIE-Gan}, U-Shape Transformer \cite{Ushape}, SS-UIE \cite{SSuie}, CE-VAE \cite{CE-VAE}, WaterMamba \cite{water-mamba}, Semi-UIR \cite{Semi-UIR}, PUIE-Net \cite{PUIE-Net}, WF-Diff \cite{WF-diff} and NU2Net \cite{NU2Net}. Wherever possible, the results for these methods have been obtained using their publicly available codes.

\subsection{Qualitative Analysis}
Figures \ref{fig:lsui} and \ref{fig:uieb} show visual comparison between Hero-Mamba and competing models. We select 3 images from both UIEB and LSUI datasets to cover a wide range of scenes. It is observed in both figures that Hero-Mamba outputs are the closest to the reference images in both datasets, with high contrast, minimal blurriness, natural-looking colors and high-fidelity local details. 

\subsection{Quantitative Analysis}
We conducted comprehensive quantitative comparison of our proposed Hero-Mamba against 10 state-of-the-art (SOTA) methods, with the results detailed in Table 1 and Table 2. As shown in Table 1, on the LSUI dataset, our Hero-Mamba model outperforms all other competing methods across all four evaluation metrics. Notably, Hero-Mamba achieves an SSIM of 0.913 and a PSNR of 25.802, significantly surpassing the second-best method, WaterMamba, which scored 0.877 and 23.463, respectively. Furthermore, our model obtains the best (lowest) LPIPS score of 0.117 and the highest FSIM score of 0.958, confirming that its enhanced images are not only mathematically accurate but also perceptually closer to the ground truth.

On the UIEB dataset, presented in Table 2, our method achieves the highest SSIM score of 0.942, outperforming all competing methods. While WF-Diff achieves a higher PSNR and NU2Net a lower LPIPS, our Hero-Mamba maintains strong performance across all metrics, positioning it as a competitive solution.

To assess our model's generalization abilities, we perform a cross-dataset evaluation, as shown in Table 3. When trained on the LSUI dataset and tested on UIEB, the model achieves a strong SSIM of 0.868 and FSIM of 0.934. Similarly, when trained on UIEB and tested on LSUI, it achieves an SSIM of 0.846 and PSNR of 20.380. These results demonstrate the robust generalization capability of Hero-Mamba.

\subsection{Ablation Study}
To demonstrate the effectiveness of each component in our proposed Hero-Mamba, we conducted a series of ablation studies with the UIEB dataset. The quantitative results of progressively adding each module are presented in Table \ref{tab:abalation}.

We start with a "Base" model, a simple encoder-decoder network with skip connections, which yields a baseline SSIM of 0.847. By incorporating the MS-Fusion block, the performance improves to 0.872, which demonstrates the effectiveness of processing features at multiple scales to reconstruct details. Subsequently, adding the Mamba-based SS2D blocks further boosts the SSIM to 0.890, validating the importance of global receptive fields and capturing long-range dependencies.

A performance jump to 0.914 is observed when we introduce FFT input, proving that our parallel, dual-domain learning strategy is effective for decoupling degradation factors. Finally, the full Hero-Mamba model, with the addition of ColorFusion block, achieves the highest SSIM of 0.942. This highlights the role of the background light prior in achieving accurate color restoration.


\section{Conclusion}
In this paper, we proposed Hero-Mamba, a novel dual-domain network for underwater image enhancement that leverages the linear complexity and long-range modeling capabilities of Mamba-based state space models. Our core contribution is a parallel architecture that processes both spatial (RGB) and spectral (FFT) features from the initial input. This dual-domain approach, powered by SS2D blocks, allows the network to efficiently model global dependencies and decouple complex degradation factors—addressing color, contrast, and detail simultaneously. Furthermore, we utilize a physics-guided ColorFusion block to accurately restore color by incorporating a background light prior, and an MS-Fusion block to reconstruct fine-grained local details. Extensive quantitative and qualitative experiments on the UIEB and LSUI datasets validate that Hero-Mamba achieves state-of-the-art performance, outperforming previous methods in key metrics like SSIM and PSNR.

\bibliography{aaai2026.bib}

\end{document}